\documentclass[11pt]{article}

\IfFileExists{acl.sty}{
\usepackage{acl}

}{
  \usepackage[margin=1in]{geometry}
  \usepackage{natbib}
}

\usepackage{times}
\usepackage{latexsym}
\usepackage[T1]{fontenc}
\usepackage[utf8]{inputenc}
\usepackage{microtype}
\usepackage{inconsolata}
\usepackage{booktabs}
\usepackage{tabularx}
\usepackage{array}
\usepackage{xcolor}
\usepackage{url}
\usepackage{comment}

\usepackage{tikz}
\usetikzlibrary{arrows.meta,positioning,fit}

\newcolumntype{Y}{>{\raggedright\arraybackslash}X}

\title{Why Current XAI Is Not Enough for Arabic NLP:\\A Critical Survey of the Explainability Gap}

\author{\textbf{Salima Lamsiyah}\textsuperscript{1}, \textbf{Ruslan Mitkov}\textsuperscript{2}\\
\textsuperscript{1} University of Luxembourg, Luxembourg \\
\textsuperscript{2} University of Alicante, Spain \\
\normalsize\texttt{salima.lamsiyah@uni.lu}
}

\begin{document}
\maketitle

\begin{abstract}
Explainable AI (XAI) is now a major theme in NLP; however, Arabic NLP remains under-explained in three connected senses. First, there is a \emph{method gap}: Arabic XAI relies heavily on a small set of post-hoc techniques such as LIME, SHAP, attention visualization, and saliency, while broader NLP XAI offers richer diagnostic, counterfactual, probing, rationale-based, and human-centered methods. Second, there is a \emph{task gap}: existing Arabic XAI work is concentrated in classification tasks, especially sentiment analysis, hate/offensive language detection, fake news, and spam, with weaker coverage of generation, retrieval, translation, summarization, structured prediction, and dialogue. Third, there is a \emph{linguistic gap}: many explanations identify influential tokens, but rarely explain Arabic-specific phenomena such as morphology, clitics, dialectal variation, diglossia, orthographic ambiguity, diacritics, code-switching, named entities, cultural references, or Classical and religious registers. This critical structured survey synthesizes the reviewed literature on Arabic XAI across text, speech, and multimodal settings. We argue that Arabic NLP does not only need explanations of model decisions; it needs explanations that are faithful to Arabic as a linguistic, cultural, and sociotechnical object. We introduce a taxonomy of tasks, methods, linguistic units, varieties, goals, and evaluation practices, and propose a research agenda for linguistically grounded Arabic XAI.
\end{abstract}

\section{Introduction}

Arabic NLP has advanced rapidly through Arabic-specific transformer models, pretrained multilingual and monolingual encoders, and Arabic-centric large language models (LLMs) \citep{abdul2021arbert,inoue2021interplay,khader2025adapting,sengupta2023jais,huang2024acegpt}. However, the same developments that improve performance also make system behavior harder to inspect, a concern widely discussed in explainable NLP and LLM explainability research \citep{zini2022explainability,luo2024local,zhao2024explainability}. This opacity is not merely a generic concern inherited from NLP. Arabic is morphologically rich, diglossic, dialectally diverse, and orthographically variable, with challenges involving clitics, segmentation, undiacritized text, dialectal variation, and register-specific usage \citep{habash2010introduction,darwish2021panoramic,inoue2021interplay}. A model may rely on a clitic, subword fragment, dialect marker, missing diacritic, named entity boundary, Qur'anic expression, or culturally loaded phrase in ways that are difficult to interpret from surface tokens alone \citep{abdelali2022post,mustafa2024interpreting,sahyoun2023aradiawer,younes2026deformar}. Explanations that are adequate for one language, register, or task may therefore be misleading when applied unchanged to Arabic.

General NLP surveys describe a broad XAI landscape, including local interpretation, rationales, probing, counterfactuals, saliency, faithfulness evaluation, and LLM explanation \citep{zini2022explainability,luo2024local,hassan2025explainable,zhao2024explainability}. Arabic-specific survey work has reviewed Arabic sentiment analysis \citep{shi2025comprehensive} and explainability for Arabic sentiment analysis \citep{alsehaimi2025toward}. However, the provided Arabic XAI literature reveals a broader \emph{explainability gap}: Arabic systems are increasingly explained, but the explanations often remain methodologically narrow, task-concentrated, and weakly connected to Arabic linguistic and socio-cultural structure.

We unpack this gap along three dimensions. The \textbf{method gap} is that many Arabic studies use LIME, SHAP, attention visualization, or saliency-style evidence as the main explanation device \citep{abdelwahab2022justifying,awadallah2022investigation,mustafa2024interpreting,aljrees2024improving,sweidan2025explainable,bouke2026novel}. These methods are valuable, but they represent only a subset of XAI for NLP. The \textbf{task gap} is that Arabic XAI is concentrated in classification: sentiment analysis, hate/offensive language, fake news, spam, reviews, news, authorship, and related social-media tasks \citep{atabuzzaman2023arabic,alwateer2025interpretable,mahouachi2026hybrid,ibrahim2024novel,alansari2026multi,alshammari2025robustness}. The \textbf{linguistic gap} is that many explanations highlight tokens or features without explaining Arabic-specific evidence: morphology, clitics, dialectal variation, diglossia, diacritics, code-switching, orthographic ambiguity, named entities, cultural references, or Classical and religious registers. A smaller set of studies points toward a richer agenda by probing Arabic transformers for morphology and dialectal information \citep{abdelali2022post}, designing explainable metrics for dialectal automatic speech recognition (ASR) \citep{sahyoun2023aradiawer}, developing diagnostic visual analytics for Arabic named entity recognition (NER) \citep{younes2026deformar}, and using knowledge-graph diagnostics for Arabic machine reading comprehension (MRC) hallucinations \citep{alghamdi2026knowledge}.

The central argument of this survey is therefore stronger than the claim that Arabic NLP needs more XAI. Arabic NLP does not only need explanations of model decisions; it needs explanations that are faithful to Arabic as a linguistic, cultural, and sociotechnical object. A heatmap over words may be a useful start, but it does not by itself explain whether a model used a dialect cue, an inflectional pattern, a named entity, a religious reference, a moderation norm, or a spurious dataset artifact.

This paper makes four contributions, structured as follows. First, it frames Arabic XAI through the method, task, and linguistic gaps. Second, it proposes a four-level account of what explanations should capture in Arabic NLP: prediction-level, model-level, linguistic, and socio-cultural evidence. Third, it organizes the surveyed literature into a critical taxonomy and a task-method coverage table that distinguish current practice from what is missing. Fourth, it proposes a concrete agenda for linguistically grounded, faithful, useful, and reproducible Arabic XAI.

\section{What Should an Explanation Explain in Arabic NLP?}
The phrase ``explainable Arabic NLP'' does not refer to a single explanatory target but encompasses several distinct ones. We propose a four-level distinction, synthesized from the reviewed literature, with the aim of rendering evaluation criteria more precise and tractable.

\paragraph{Prediction-level explanation.}
At the most fundamental level, an explanation accounts for why a model assigns a specific label, returns a particular score, retrieves a given result, or generates a specific output. Most Arabic XAI classification papers operate at this level, using local feature attribution to identify influential words or features for sentiment, fake news, spam, or hate-speech decisions \citep{abdelwahab2022justifying,awadallah2022investigation,aljrees2024improving,ibrahim2024novel,bouke2026novel,alwateer2025interpretable}. Prediction-level explanations can support debugging and user trust, but they are not automatically faithful or linguistically meaningful.

\paragraph{Model-level explanation.}
A model-level explanation addresses the internal representations, layers, features, or mechanisms that the model employs. This level is especially relevant for Arabic transformers and multilingual encoders, where behavior may be distributed across layers and subword representations. \citet{abdelali2022post}, for example, probe Arabic transformer models for morphology, syntax, and dialect identification. Attention-based sentiment architectures \citep{berrimi2023attention,ghasemi2023deep,sweidan2025explainable} and LLM adaptation studies \citep{khader2025adapting} are also relevant, although attention or benchmark performance alone does not settle whether the explanation is faithful.

\paragraph{Linguistic explanation.}
A linguistic explanation identifies which Arabic linguistic features -- morphological, syntactic, semantic, dialectal, or orthographic -- are operative in the model's predictions.  This level is where the Arabic explainability gap is most visible. For Arabic, the relevant unit may be a stem, clitic, lemma, diacritic, named entity, multiword expression, dialect phrase, or orthographic variant rather than a whitespace-delimited token. Work on dialectal ASR evaluation \citep{sahyoun2023aradiawer}, Arabic NER diagnostics \citep{younes2026deformar}, Arabic readability \citep{rabih2025interpretable}, and transformer probing \citep{abdelali2022post} begins to connect explanations to linguistic structure, but this remains less common than token attribution.

\paragraph{Socio-cultural explanation.}
A socio-cultural explanation addresses how a model's interpretation is shaped by cultural context, moderation norms, religious or classical references, identity terms, user expectations, and the broader social meaning of language use. This level is essential for harmful-content moderation, propaganda, Qur'anic semantic search, and high-stakes applications. User-centered Arabic hate-speech explanation work \citep{al2026explaining}, neuro-symbolic offensive-language detection \citep{mahouachi2026hybrid}, propagandistic meme explanation \citep{kmainasi2025memeintel}, Qur'anic semantic search \citep{mustafa2024interpreting}, and knowledge-graph diagnostics for Qur'anic MRC \citep{alghamdi2026knowledge} illustrate why Arabic XAI cannot be reduced to visually plausible word highlights.

\section{A Critical Taxonomy}

Table~\ref{tab:taxonomy} summarizes the principal dimensions of the explainability gap, distinguishing current practice from missing explanatory capacity and grounding each absence in an Arabic-specific motivation.

The taxonomy demonstrates that the Arabic XAI gap cannot be reduced to a shortage of explanation papers alone. Rather, the same pattern recurs across several axes: explanations are often attached to already trained systems as post-hoc artifacts, while the Arabic-specific object of explanation remains underspecified. For example, LIME, SHAP, attention visualization, and feature-importance methods can identify influential surface units in sentiment, fake-news, spam, or hate-speech classifiers \citep{abdelwahab2022justifying,awadallah2022investigation,aljrees2024improving,bouke2026novel}, but they do not by themselves establish whether the model has used a clitic, a stem, a dialectal marker, a diacritic-sensitive ambiguity, or a named-entity boundary in a linguistically meaningful way. In this sense, the method, task, and linguistic-unit axes are tightly coupled: when the task is framed as label prediction and the method is framed as token attribution, the resulting explanation is likely to remain at the prediction level even when the underlying error is morphological, dialectal, semantic, or socio-cultural.

The table also makes clear that Arabic XAI requires a stronger and more explicit alignment between explanation goals and evaluation practice.  Some reviewed work already moves beyond generic attribution through probing, visual analytics, explainable ASR metrics, knowledge-graph diagnostics, neuro-symbolic modeling, and user-centered evaluation \citep{abdelali2022post,sahyoun2023aradiawer,younes2026deformar,alghamdi2026knowledge,mahouachi2026hybrid,al2026explaining}. These studies are important because they make the explanatory target more explicit: representations, error types, entity boundaries, semantic triples, rules, or user informativeness. However, they remain scattered rather than forming a shared protocol. The central challenge, therefore, is not to replace post-hoc methods wholesale, but to embed them in Arabic-aware evaluation designs that test faithfulness, stability, usefulness, and fairness under the linguistic variation that Arabic NLP systems actually encounter.

\begin{table*}[h!]
\centering
\scriptsize
\begin{tabularx}{\textwidth}{p{0.08\textwidth}p{0.22\textwidth}p{0.18\textwidth}p{0.2\textwidth}Y}
\toprule
\textbf{Axis} & \textbf{Current practice} & \textbf{Representative works} & \textbf{What is missing} & \textbf{Why it matters for Arabic NLP} \\
\midrule
Methods & Local post-hoc explanations dominate: LIME, SHAP, attention visualization, saliency, and feature importance. Some work uses probing, visual analytics, diagnostic metrics, knowledge graphs, rationales, or neuro-symbolic rules. & \citet{abdelwahab2022justifying}, \citet{awadallah2022investigation}, \citet{azzem2024exploring}, \citet{mustafa2024interpreting}, \citet{abdelali2022post}, \citet{younes2026deformar}, \citet{alghamdi2026knowledge}, \citet{mahouachi2026hybrid} & Counterfactuals, causal tests, Arabic-preserving perturbations, faithful rationale evaluation, mechanistic analysis, and systematic human evaluation are rare in the reviewed Arabic-specific work. & Arabic tokenization, morphology, spelling, and dialect variation can make word-level attributions unstable or linguistically misleading. \\
\addlinespace
NLP Tasks & Classification tasks dominate, especially sentiment, hate/offensive language, fake news, spam, app reviews, news classification, authorship, and mental health detection. & \citet{abdelwahab2022justifying}, \citet{atabuzzaman2023arabic}, \citet{alwateer2025interpretable}, \citet{aljrees2024improving}, \citet{ibrahim2024novel}, \citet{bouke2026novel}, \citet{alansari2026multi}, \citet{kumar2023explainable} & Less coverage for generation, QA, retrieval-augmented generation, summarization, translation, dialogue, parsing, grammatical error correction, and information extraction. & High-stakes Arabic systems increasingly involve open-ended generation, grounding, retrieval, and structured decisions where label-level explanations are insufficient. \\
\addlinespace
Linguistic units & Most explanations are token-, word-, or feature-level. A smaller set studies layers, neurons, ASR error types, NER components, readability features, or semantic triples. & \citet{mustafa2024interpreting}, \citet{abdelali2022post}, \citet{sahyoun2023aradiawer}, \citet{younes2026deformar}, \citet{rabih2025interpretable}, \citet{alghamdi2026knowledge} & Explanations rarely target morphemes, clitics, stems, lemmas, diacritics, multiword expressions, named-entity boundaries, or dialectal phrases. & Arabic linguistic evidence is often below or above the whitespace-token level; explanations that ignore this can explain the wrong object. \\
\addlinespace
Varieties and registers & Studies cover MSA, social-media Arabic, dialectal Arabic, Classical/Qur'anic Arabic, and multimodal Arabic-language settings, but variety is often metadata rather than an evaluated explanatory factor. & \citet{awadallah2022investigation}, \citet{sahyoun2023aradiawer}, \citet{abdelali2022post}, \citet{mustafa2024interpreting}, \citet{kboubi2025fine}, \citet{kmainasi2025memeintel} & Cross-variety explanation evaluation for MSA, regional dialects, Arabizi, code-switching, Classical Arabic, and social-media spelling variation is limited. & A model may rely on dialect, register, or orthographic cues rather than task-relevant semantics, creating fairness and robustness risks. \\
\addlinespace
Goals & Explanations are used for trust, debugging, model comparison, user informativeness, safer moderation, hallucination diagnosis, and linguistic analysis. & \citet{al2026explaining}, \citet{younes2026deformar}, \citet{alghamdi2026knowledge}, \citet{abdelali2022post}, \citet{mahouachi2026hybrid} & Papers often blur explanation for users, developers, linguists, domain experts, and affected communities. & Arabic applications involve different stakeholders: annotators, moderators, clinicians, educators, religious-domain users, and dialect communities may need different explanations. \\
\addlinespace
Evaluation & Many papers report predictive metrics and selected explanation examples. Some compare XAI techniques, include user-centered evaluation, or provide diagnostic tools. & \citet{azzem2024exploring}, \citet{al2026explaining}, \citet{kmainasi2025memeintel}, \citet{bouke2026novel}, \citet{younes2026deformar} & Faithfulness, plausibility, stability, usefulness, fairness, and reproducibility protocols are not standardized. & Without Arabic-aware evaluation, visually plausible explanations may hide brittle behavior under normalization, segmentation, diacritic, or dialect changes. \\
\bottomrule
\end{tabularx}
\caption{Critical taxonomy of the Arabic XAI explainability gap.}
\label{tab:taxonomy}
\end{table*}

\section{Task-Method Coverage}
Table~\ref{tab:coverage} maps representative tasks to explanation methods and gaps. The asymmetry is clear: sentiment analysis and harmful-content classification form the strongest clusters, whereas generation, retrieval-augmented systems, translation, summarization, parsing, dialogue, and structured linguistic analysis remain weakly covered in Arabic XAI. The table also shows that method coverage is narrower than task coverage: many tasks rely on the same small family of post-hoc explainers.

\begin{table*}[h!]
\centering
\scriptsize
\begin{tabularx}{\textwidth}{p{0.1\textwidth}p{0.13\textwidth}p{0.22\textwidth}p{0.24\textwidth}Y}
\toprule
\textbf{Task cluster} & \textbf{Coverage} & \textbf{Methods: used vs. missing} & \textbf{Variety and evaluation gaps} & \textbf{Representative works} \\
\midrule
Sentiment, emotion, opinions & Well covered relative to other Arabic XAI tasks & LIME, SHAP, attention, local surrogates, empirical method comparison, and interfaces are common; underused: counterfactuals, Arabic-preserving perturbations, rationale benchmarks. & MSA, social-media Arabic, and some dialectal data appear, but negation, sarcasm, dialect morphology, and diacritics are rarely evaluation variables. Faithfulness and stability are not consistently tested. & \citet{abdelwahab2022justifying}, \citet{awadallah2022investigation}, \citet{atabuzzaman2023arabic}, \citet{azzem2024exploring}, \citet{alsehaimi2025toward}, \citet{sweidan2025explainable}, \citet{chafiqui2024xs2a} \\
\addlinespace
Hate, offensive language, propaganda & Moderately covered and socially important & LLM-based interpretability, user-facing explanations, neuro-symbolic/fuzzy rules, and multimodal rationales appear; underused: fairness audits, dialect-stratified explanation tests, moderation-specific rationale evaluation. & Dialectal and social-media Arabic are central, but cross-dialect, identity-term, target-group, and cultural-context evaluation remains limited. & \citet{alwateer2025interpretable}, \citet{al2026explaining}, \citet{mahouachi2026hybrid}, \citet{kmainasi2025memeintel} \\
\addlinespace
Fake news and spam & Narrow but visible & LIME, SHAP, feature importance, and ensemble attribution are used; underused: evidence-chain explanations, claim/source grounding, temporal and entity-aware analysis. & Arabic news and social-media settings are used, but evaluation often stops at salient features and predictive scores rather than reasoning about misinformation mechanisms. & \citet{aljrees2024improving}, \citet{ibrahim2024novel}, \citet{bouke2026novel} \\
\addlinespace
Semantic search, MRC, QA, hallucination & Emerging & LIME/SHAP for retrieval and knowledge-graph triple diagnostics for hallucination; underused: answer-grounding faithfulness, citation-level evidence, retrieval-error attribution. & Classical/Qur'anic Arabic is represented in specialized settings, but broader Arabic QA and RAG evaluation remains limited. & \citet{mustafa2024interpreting}, \citet{alghamdi2026knowledge} \\
\addlinespace
NER, ASR, readability, dialect ID & Emerging and more linguistically grounded & Probing, visual analytics, explainable ASR metrics, and interpretable linguistic features; underused: shared explanation benchmarks and cross-task diagnostic protocols. & These works better target morphology, dialect, errors, entities, and readability, but they remain isolated and not yet a common Arabic XAI paradigm. & \citet{abdelali2022post}, \citet{sahyoun2023aradiawer}, \citet{younes2026deformar}, \citet{rabih2025interpretable}, \citet{kboubi2025fine} \\
\addlinespace
Reviews, news, authorship, mental health & Applied but uneven & Attention, LIME/SHAP, stylometric features, and robustness analysis; underused: expert-centered evaluation and harm-aware explanation protocols. & These settings can be high-stakes, but user utility and linguistic robustness are often less developed than classification accuracy. & \citet{hossain2024hybrid}, \citet{alansari2026multi}, \citet{alshammari2025robustness}, \citet{kumar2023explainable} \\
\addlinespace
Multimodal Arabic-language tasks & Growing but adjacent & Grad-CAM, interpretable visual concepts, multimodal rationales; underused: joint evaluation of visual grounding, Arabic language quality, and explanation faithfulness. & Arabic sign language, captions, and memes expand the field beyond text, but explanation quality is harder to separate from vision and generation quality. & \citet{baghdadi2024toward}, \citet{balat2025revolutionizing}, \citet{elchafei2025multimodal}, \citet{kmainasi2025memeintel} \\
\addlinespace
Generation, machine translation, summarization, dialogue, RAG & Weakly covered in the literature & Adjacent LLM adaptation and linguistic benchmark work exist; underused: hallucination explanations, retrieval attribution, translation-error explanations, summary faithfulness explanations. & These tasks are central to modern Arabic NLP, but explanations must account for grounding, fluency, factuality, morphology, and discourse rather than a single class label. & \citet{khader2025adapting}, \citet{zbib2026aralingbench}, \citet{alghamdi2026knowledge} \\
\bottomrule
\end{tabularx}
\caption{Task-method coverage and asymmetries in the surveyed Arabic XAI literature.}
\label{tab:coverage}
\end{table*}

\subsection{Sentiment and Opinion Mining: Useful but Over-Represented}

Arabic sentiment analysis is the clearest default testbed for XAI. Studies combine sentiment classifiers with LIME-style local explanations for domain-specific medical opinions \citep{abdelwahab2022justifying}, multi-dialect sentiment classification \citep{awadallah2022investigation}, and noisy deep explainable sentiment modeling \citep{atabuzzaman2023arabic}. Attention-based sentiment architectures and cross-lingual sentiment models use attention or contextual representations as interpretive evidence \citep{berrimi2023attention,ghasemi2023deep}, while newer work adds SHAP, multi-self-attention, empirical comparison of explanation techniques, or platform-level explanation interfaces \citep{azzem2024exploring,chafiqui2024xs2a,sweidan2025explainable}.

These works contribute an important baseline: they make Arabic classifier decisions inspectable and show that XAI can be integrated into practical sentiment pipelines. However, much of this work remains prediction-level. It often explains which words influenced a label, not whether the model captured negation, dialectal intensification, sarcasm, stance, topic confounds, or morphology. This distinction matters because a visually plausible explanation can be useful for demonstration while still failing faithfulness, stability, or Arabic-awareness. The sentiment scoping review by \citet{alsehaimi2025toward} is therefore important as a consolidation of the subfield, but it also underscores the limits of sentiment-centered Arabic XAI as a proxy for the whole language.

\subsection{Harmful Content: Explanations Need Social Context}

Hate speech, offensive language, and propaganda expose the socio-cultural level of explanation. \citet{alwateer2025interpretable} study interpretable Arabic hate-speech detection with LLMs; \citet{al2026explaining} evaluate perceived informativeness of XAI representations for Arabic hate-speech detection with Arabic-speaking users; \citet{mahouachi2026hybrid} uses a hybrid multi-view and neuro-symbolic approach for Arabic offensive-language detection; and \citet{kmainasi2025memeintel} study explanation-enhanced detection of propagandistic Arabic memes.

Compared with sentiment work, this cluster more clearly motivates user-centered and safety-oriented explanations. It also shows why token salience is insufficient. A word can be harmful, quoted, reclaimed, dialect-specific, target-dependent, or embedded in a meme. The contribution of this line of work is to connect explainability with moderation and social meaning. Its limitation is that the field still lacks shared protocols for testing whether explanations remain useful and fair across dialects, regions, identity terms, targets, and cultural references. In our synthesis, harmful-content XAI is one of the strongest arguments for treating Arabic as a sociotechnical object rather than a string of tokens.

\subsection{Information Integrity: Attribution Is Not Evidence}

Arabic fake-news and spam studies use explainability to inspect high-performing classifiers. \citet{aljrees2024improving} combines LIME with an ELMo-based tri-ensemble fake-news model; \citet{ibrahim2024novel} use large-language-feature embeddings, CNN-LSTM ensembles, and explainable AI for Arabic fake-news classification; and \citet{bouke2026novel} integrates LIME and SHAP into an Arabic spam-detection pipeline based on LightGBM.

These studies are useful for feature inspection and model debugging. The critical issue is that attribution is not the same as evidence. A highlighted token may reveal what the classifier used, but not whether the model identified deception, propaganda framing, source credibility, temporal inconsistency, or a culturally grounded claim. For Arabic misinformation, explanations should ideally connect entities, claims, sources, and context. The reviewed work moves toward transparency, but it rarely explains the information-integrity phenomenon itself.

\subsection{Diagnostics Beyond Classification}

Work that most substantively engages with Arabic-specific phenomena frequently falls outside the conventions of standard post-hoc explanation. \citet{mustafa2024interpreting} use LIME and SHAP for Arabic transformer models in Qur'anic semantic search, where explanation must confront high-register and religious text. \citet{alghamdi2026knowledge} use knowledge-graph triple diagnostics for Arabic MRC hallucinations, shifting from word salience to relational grounding. \citet{abdelali2022post} probes Arabic transformers for morphology, syntax, and dialect information. \citet{sahyoun2023aradiawer} propose an explainable ASR metric for dialectal Arabic, and \citet{younes2026deformar} develops visual analytics for Arabic NER evaluation. \citet{rabih2025interpretable} studies interpretable measures for Arabic readability.

This group shows what linguistically grounded Arabic XAI can look like. Rather than asking only which token affected a label, these works ask where linguistic information is encoded, what kind of recognition error occurred, which entity boundaries or annotations are problematic, which readability features matter, or whether a generated answer is semantically grounded. Their limitation is scale and standardization: they are promising case studies and frameworks, not yet a shared paradigm for Arabic XAI evaluation.

\subsection{Multimodal and LLM-Adjacent Work}

Arabic XAI is also expanding beyond text-only classification. Arabic sign language recognition uses LIME or Grad-CAM-style visual explanations \citep{baghdadi2024toward,balat2025revolutionizing}; Arabic captioning uses interpretable visual concept integration \citep{elchafei2025multimodal}; and meme detection connects multimodal classification with rationale-oriented resources \citep{kmainasi2025memeintel}. LLM-oriented Arabic work discusses adaptation, evaluation, and ethical deployment \citep{khader2025adapting}, while AraLingBench evaluates Arabic linguistic capabilities of LLMs \citep{zbib2026aralingbench}.

These works widen the scope of Arabic XAI, but they also make evaluation harder. For multimodal tasks, explanation quality must be separated from visual grounding and Arabic generation quality. For LLMs, explanation must be connected to factual grounding, linguistic competence, hallucination, and user expectations. Within the reviewed literature, Arabic LLM explainability remains more of an emerging requirement than a mature methodology.

\section{Critical Discussion: The existing Gaps}

\subsection{The Method Gap}

The reviewed literature shows a clear preference for methods that are easy to attach to existing classifiers: LIME, SHAP, attention visualization, saliency, or feature importance. This is understandable because model-agnostic methods are accessible and produce intuitive visual artifacts. The problem is that these artifacts are often treated as explanations without enough evidence that they are faithful, stable, or aligned with Arabic linguistic units. General NLP XAI surveys emphasize that local explanations, rationales, probing, counterfactuals, and faithfulness tests answer different questions \citep{zini2022explainability,luo2024local,hassan2025explainable}. In the Arabic corpus, richer methods appear in isolated forms, including probing \citep{abdelali2022post}, visual analytics \citep{younes2026deformar}, knowledge-graph diagnostics \citep{alghamdi2026knowledge}, user-centered evaluation \citep{al2026explaining}, and neuro-symbolic modeling \citep{mahouachi2026hybrid}, but they have not yet become common practice.

\subsection{The Task Gap}

Arabic XAI is disproportionately shaped by classification. Sentiment analysis, hate/offensive language, fake news, spam, app reviews, news classification, authorship, and mental health detection supply many of the examples and methods \citep{abdelwahab2022justifying,atabuzzaman2023arabic,alwateer2025interpretable,aljrees2024improving,bouke2026novel,alansari2026multi,kumar2023explainable,alshammari2025robustness}. Classification is a useful starting point, but it encourages explanations that identify label-supporting tokens. Modern Arabic NLP increasingly includes retrieval, generation, machine reading comprehension, captioning, LLM adaptation, and linguistic benchmarking \citep{mustafa2024interpreting,alghamdi2026knowledge,elchafei2025multimodal,khader2025adapting,zbib2026aralingbench}. These tasks require explanations of grounding, hallucination, alignment, fluency, factuality, entity tracking, and discourse, not only explanations of class labels.

\subsection{The Linguistic Gap}

We argue that the most important gap is linguistic. Many Arabic XAI papers explain which input words influenced a prediction, but not what Arabic evidence the model used. This is not a minor technical issue. Arabic preprocessing choices, including normalization, segmentation, stop-word removal, dialect filtering, subword tokenization, and diacritic handling, can change the unit being explained. If an explanation highlights a whitespace token while the model operates over subwords, or if preprocessing removes clitics or diacritics, the explanation may be difficult to interpret linguistically. Work on transformer probing, dialectal ASR, NER diagnostics, readability, and Qur'anic MRC demonstrates that explanations can be tied to morphology, dialect, recognition errors, entity boundaries, readability features, and semantic relations \citep{abdelali2022post,sahyoun2023aradiawer,younes2026deformar,rabih2025interpretable,alghamdi2026knowledge}. This is the direction in which Arabic XAI should move.

\subsection{The Evaluation Gap}

Faithfulness, plausibility, stability, usefulness, and fairness are distinct evaluation targets \citep{zini2022explainability,luo2024local}. Arabic XAI papers increasingly acknowledge these criteria, and some compare explanation techniques or evaluate user informativeness \citep{azzem2024exploring,al2026explaining}. However, many studies still rely on predictive accuracy plus selected explanation examples. For Arabic, this is especially weak because an explanation can change under spelling variation, normalization, segmentation, dialectal paraphrase, or diacritic removal while the meaning remains similar. In high-stakes settings--health sentiment \citep{abdelwahab2022justifying,sweidan2025explainable}, mental health detection \citep{kumar2023explainable}, hate speech \citep{alwateer2025interpretable,al2026explaining}, fake news \citep{aljrees2024improving,ibrahim2024novel}, Qur'anic search and MRC \citep{mustafa2024interpreting,alghamdi2026knowledge}, and Arabic sign language recognition \citep{baghdadi2024toward,balat2025revolutionizing}--explanations should be evaluated for decision support, error discovery, harm reduction, and stakeholder usefulness, not only visual plausibility.

\section{Toward Linguistically Grounded Arabic XAI}
A linguistically grounded agenda for Arabic XAI should move beyond adding generic explanation methods to Arabic models. It should instead redesign explanation targets, evaluation protocols, and reporting standards around the linguistic and sociotechnical properties of Arabic itself. The following directions outline how future work can make explanations more faithful to model behavior, more stable under Arabic-specific variation, more useful to Arabic-speaking users and domain experts, and more reproducible across tasks, datasets, and varieties.

\paragraph{Arabic-aware units of explanation.}
Explanations should be evaluated at levels that are meaningful for Arabic: morphemes, clitics, stems, lemmas, diacritics, named entities, multiword expressions, dialectal phrases, and discourse cues. Token-level heatmaps are insufficient when model decisions depend on morphology or orthography. Future papers should report the relationship between the model's internal tokenization and the units shown to users.

\paragraph{Arabic-preserving perturbation tests.}
Arabic XAI needs perturbation tests that preserve meaning while varying surface form: spelling variation, normalization, clitic segmentation, dialectal paraphrases, and diacritic restoration or removal. Such tests would help distinguish robust explanations from artifacts of preprocessing or tokenization. This would extend the faithfulness and stability concerns emphasized in general NLP XAI \citep{zini2022explainability,luo2024local} to Arabic-specific forms.

\paragraph{Cross-variety explanation evaluation.}
Benchmarks should evaluate explanations across MSA, regional dialects, Classical Arabic, social-media Arabic, Arabizi, and code-switched Arabic. Work on dialectal ASR and Arabic transformer probing shows that variety can be studied systematically \citep{sahyoun2023aradiawer,abdelali2022post}; Arabic XAI should make variety an explanation variable, not merely a dataset description.

\paragraph{Human rationale benchmarks with Arabic expertise.}
Arabic XAI needs shared datasets with human rationales, expert annotations, and task-specific explanation quality labels. User-centered work on Arabic hate-speech explanation \citep{al2026explaining} and explanation-enhanced multimodal resources \citep{kmainasi2025memeintel} point to this direction. Annotators should include Arabic speakers with dialectal competence and, where necessary, domain expertise in moderation, health, education, religious text, or accessibility.

\paragraph{Explanation evaluation for Arabic LLMs, RAG, and generation.}
Future work should target Arabic LLMs, retrieval-augmented generation, QA, summarization, machine translation, dialogue, and hallucination detection. Arabic LLM adaptation and linguistic benchmarking are already emerging \citep{khader2025adapting,zbib2026aralingbench}, and knowledge-graph diagnostics show one path for hallucination analysis \citep{alghamdi2026knowledge}. Explanations for these systems should address factual grounding, retrieval evidence, morphology, syntax, discourse, and answer-source alignment.

\paragraph{Faithfulness, plausibility, stability, usefulness, and fairness protocols.}
Arabic XAI needs evaluation protocols that separate what the model actually used from what users find plausible. Faithfulness tests, human plausibility judgments, stability under Arabic-preserving perturbations, usefulness for stakeholders, and fairness across varieties and identity terms should be reported separately. This distinction is especially important in moderation, health, education, religious-domain applications, and information integrity.

\paragraph{Reproducibility standards.}
Arabic XAI papers should report dataset variety, preprocessing, normalization, tokenization, segmentation, diacritic handling, model checkpoints, explanation parameters, evaluation metrics, and known validity limits. Where possible, studies should release code, trained models or inference scripts, explanation outputs, rationale annotations, and diagnostic error analyses. Without these details, it is difficult to know whether a method is better for Arabic or simply easier to visualize.

\section{Conclusion}

Arabic XAI is an emerging and necessary research direction, but the field is still shaped by a clear explainability gap. Current work has demonstrated the value of LIME, SHAP, attention visualization, probing, visual analytics, knowledge graphs, neuro-symbolic methods, and user-centered explanations. However, the dominant pattern remains output-level explanation for classification tasks, especially sentiment analysis, harmful-content detection, fake news, and spam. This leaves important parts of Arabic NLP under-explained, including generation, retrieval, QA, summarization, translation, structured prediction, and LLM-based applications.

The main message of this survey is that Arabic NLP does not only need explanations of model decisions; it needs explanations that are faithful to Arabic as a linguistic, cultural, and sociotechnical object. Explanations should account for morphology, clitics, dialectal variation, diglossia, diacritics, orthographic ambiguity, named entities, cultural references, and Classical or religious registers. By identifying the method gap, task gap, and linguistic gap, this survey aims to provide a clearer foundation for future work in Arabic XAI. Advancing this agenda is important for building Arabic NLP systems that are not only accurate, but also transparent, trustworthy, fair, and useful in the diverse contexts where Arabic is used. Future research should therefore move from explaining predictions to explaining Arabic itself through Arabic-aware explanation units, cross-variety evaluation, perturbation testing, human rationale benchmarks, LLM/RAG explanation protocols, and stronger reproducibility standards.

\section{Survey Methodology and Scope}

This paper is a \emph{critical structured survey}, not a fully systematic review. We searched across major academic databases and indexes, including ACL Anthology, ACM Digital Library, IEEE Xplore, ScienceDirect, SpringerLink, MDPI, arXiv/preprint venues, and Google Scholar, using combinations of keywords such as Arabic NLP, explainable AI, interpretability, transparency, XAI, LIME, SHAP, attention, saliency, LLMs, sentiment analysis, hate speech, fake news, spam, ASR, NER, MRC, hallucination, readability, and multimodal Arabic NLP. General XAI-for-NLP and XAI-for-LLMs surveys are used for background and terminology \citep{zini2022explainability,luo2024local,hassan2025explainable,zhao2024explainability}.

We included studies when they met two criteria: (i) the task, dataset, model, or output was Arabic or Arabic-language adjacent; and (ii) the work explicitly addressed explanation, interpretability, transparency, diagnostic analysis, or human-facing explanation. We excluded papers that only reported Arabic NLP performance without an explicit explanatory component, and we treat attention-based models cautiously because attention is not necessarily an explanation \citep{zini2022explainability,luo2024local}. Each included work was annotated by task, model family, XAI method, Arabic variety or register, explanation goal, evaluation practice, and degree of linguistic grounding. The relatively small number of directly relevant papers confirms the lack of dedicated work on XAI for Arabic NLP and motivates framing this paper as a critical survey of an emerging, underdeveloped area.

\section*{Limitations}
This survey provides a critical synthesis of recent work on explainability for Arabic NLP, but it does not claim to be an exhaustive systematic review of all Arabic NLP papers that may contain interpretable components. Its scope is limited to studies that explicitly frame their contribution in terms of explainability, interpretability, transparency, or diagnostic analysis. As a result, some relevant work on Arabic model analysis, evaluation, or error diagnosis may fall outside the discussion if it does not use XAI terminology. 

A second limitation is that the reviewed literature is uneven across tasks and publication stages. Some areas, such as sentiment analysis and harmful-content detection, are represented more strongly than others, while emerging topics such as Arabic LLM explainability, retrieval-augmented generation, hallucination analysis, and multimodal Arabic XAI remain relatively new. Consequently, the taxonomy and agenda proposed here should be seen as a structured snapshot of a developing field rather than a fixed classification.

\section*{Use of AI Assistance}
In preparing this work, we used AI-assisted tools for language editing, such as spell checking and stylistic revisions with Grammarly and ChatGPT. The authors take full responsibility for the scientific content, analyses, and conclusions presented in this work.

\bibliography{arabic_xai_refs}

\end{document}